\documentclass[letterpaper]{article} 
\usepackage{aaai25}  
\usepackage{times}  
\usepackage{helvet}  
\usepackage{courier}  
\usepackage[hyphens]{url}  
\usepackage{graphicx} 
\usepackage{natbib}  
\usepackage{caption} 
\usepackage{algorithm}
\usepackage{algorithmic}
\usepackage{amsmath}
\usepackage{adjustbox}
\usepackage{booktabs}       
\usepackage{amsfonts}       
\usepackage{nicefrac}       
\usepackage{microtype}      
\usepackage{xcolor}         
\usepackage{amsmath}
\usepackage{amssymb}
\usepackage{amsmath}
\usepackage{mathtools}
\usepackage{multirow}
\usepackage{bm}
\usepackage{graphicx}
\usepackage{caption}
\usepackage{subcaption}

\newtheorem{definition}{Definition}

\DeclareMathOperator*{\argmax}{arg\,max}
\DeclareMathOperator*{\argmin}{arg\,min}

\newcommand\hl[1]{\textbf{#1.}}
\newcommand\la{\leftarrow}

\def\ddefloop#1{\ifx\ddefloop#1\else\ddef{#1}\expandafter\ddefloop\fi}

\def\ddef#1{\expandafter\def\csname v#1\endcsname{\ensuremath{\boldsymbol{#1}}}}
\ddefloop ABCDEFGHIJKLMNOPQRSTUVWXYZabcdefghijklmnopqrstuvwxyz\ddefloop

\def\ddef#1{\expandafter\def\csname v#1\endcsname{\ensuremath{\boldsymbol{\csname #1\endcsname}}}}
\ddefloop {alpha}{beta}{gamma}{delta}{epsilon}{varepsilon}{zeta}{eta}{theta}{vartheta}{iota}{kappa}{lambda}{mu}{nu}{xi}{pi}{varpi}{rho}{varrho}{sigma}{varsigma}{tau}{upsilon}{phi}{varphi}{chi}{psi}{omega}{Gamma}{Delta}{Theta}{Lambda}{Xi}{Pi}{Sigma}{varSigma}{Upsilon}{Phi}{Psi}{Omega}{ell}\ddefloop

\def\ddef#1{\expandafter\def\csname bb#1\endcsname{\ensuremath{\mathbb{#1}}}}
\ddefloop ABCDEFGHIJKLMNOPQRSTUVWXYZ\ddefloop

\usepackage{newfloat}
\usepackage{listings}
\DeclareCaptionStyle{ruled}{labelfont=normalfont,labelsep=colon,strut=off} 
\floatstyle{ruled}
\newfloat{listing}{tb}{lst}{}
\floatname{listing}{Listing}
\title{Improving Pareto Set Learning for Expensive Multi-objective Optimization \\via Stein Variational Hypernetworks}
\author{
    Minh-Duc Nguyen\textsuperscript{\rm 1},
    Phuong Mai Dinh\textsuperscript{\rm 1, \rm 2},
    Quang-Huy Nguyen\textsuperscript{\rm 1, \rm 3 },\\
    Long P. Hoang\textsuperscript{\rm 4},
    Dung D. Le\textsuperscript{\rm 1}
}
\affiliations{
    \textsuperscript{\rm 1} College of Engineering and Computer Science, VinUniversity, Vietnam\\
    \textsuperscript{\rm 2} VinUni-Illinois Smart Health Center, VinUniversity, Vietnam

     \textsuperscript{\rm 3} Department of Computer Science and Engineering, The Ohio State University, USA
     \\
     \textsuperscript{\rm 4} Information Systems Technology and Design, Singapore University of Technology and Design, Singapore

    duc.nm2@vinuni.edu.vn, phuong.dm2@vinuni.edu.vn, nguyen.2959@osu.edu,\\long\_hoang@mymail.sutd.edu.sg, dung.ld@vinuni.edu.vn
}

\usepackage{bibentry}

\begin{document}

\maketitle

\begin{abstract}
Expensive multi-objective optimization problems (EMOPs) are common in real-world scenarios where evaluating objective functions is costly and involves extensive computations or physical experiments. Current Pareto set learning methods for such problems often rely on surrogate models like Gaussian processes to approximate the objective functions. These surrogate models can become fragmented, resulting in numerous small uncertain regions between explored solutions. When using acquisition functions such as the Lower Confidence Bound (LCB), these uncertain regions can turn into pseudo-local optima, complicating the search for globally optimal solutions. To address these challenges, we propose a novel approach called SVH-PSL, which integrates Stein Variational Gradient Descent (SVGD) with Hypernetworks for efficient Pareto set learning. Our method addresses the issues of fragmented surrogate models and pseudo-local optima by collectively moving particles in a manner that smooths out the solution space. The particles interact with each other through a kernel function, which helps maintain diversity and encourages the exploration of underexplored regions. This kernel-based interaction prevents particles from clustering around pseudo-local optima and promotes convergence towards globally optimal solutions. Our approach aims to establish robust relationships between trade-off reference vectors and their corresponding true Pareto solutions, overcoming the limitations of existing methods. Through extensive experiments across both synthetic and real-world MOO benchmarks, we demonstrate that SVH-PSL significantly improves the quality of the learned Pareto set, offering a promising solution for expensive multi-objective optimization problems.
\end{abstract}

%

\section{Introduction}

Multi-objective optimization (MOO) has numerous essential and practical applications across various fields, from text-to-image generation \cite{lee2024parrot} to ejector design for fuel cell systems \cite{hou2024optimization}. However, real-world MOO problems often involve multiple conflicting and expensive-to-evaluate objectives. For example, recommendation systems must achieve a balance between precision and efficiency \cite{le2017indexable} or between precision and revenue \cite{milojkovic2019multi}, battery usage optimization requires trade-offs between performance and lifetime \cite{Attia2020ClosedloopOO}, robotic radiosurgery involves coordination of internal and external motion \cite{Schweikard2000RoboticMC}. Traditional MOO methods, such as hyperparameter tuning, seek a finite set of Pareto-optimal solutions representing trade-offs between objectives. However, as the number of objectives increases, approximating the Pareto front becomes exponentially costly, leading to a trade-off between generalization and computational time  \cite{swersky2013multi}.


Pareto Set Learning (PSL) \cite{navon2021learning, hoang2023improving, tuan2024framework} is a promising approach that allows users to explore the entire Pareto front for MOO problems by learning a parametric mapping to align trade-off preference weights assigned to objectives with their corresponding Pareto optimal solutions. The optimized mapping model enables real-time adjustments between objectives.

Expensive objective problems refer to a class of real-world problems in which evaluating each objective is expensive. For example, evaluating a battery's lifetime costs a newly developed battery or evaluating robotic radiosurgery costs a lot of one-time-use medical equipment. To address this challenge, researchers have developed surrogate model techniques \cite{he2023review} that estimate the actual objective functions, thus minimizing the necessity for costly evaluations.
Lin et al. \cite{lin2022pareto} is one of the pioneer teams in the study of Pareto set learning for the EMOPs by acquiring knowledge of the entire Pareto front, referred to as PSL-MOBO. This work employs the Pareto set learning with Multi-objective Bayesian Optimization (MOBO) \cite{laumanns2002bayesian} to effectively solve black- box expensive optimization problems by minimizing the number of function evaluations.

\hl{Challenge} We observed that PSL-MOBO often struggles with instability due to its tendency to become trapped in pseudo-local optima. This issue arises because the optimization process can mistakenly identify these false optima as true optimal solutions, causing the algorithm to converge prematurely and fail to explore the solution space thoroughly. As a result, PSL-MOBO might not effectively approximate the true Pareto front, especially in complex or high-dimensional spaces where numerous local optima are present.  Consequently, learning the Pareto front requires an advanced algorithm to avoid premature convergence while simultaneously utilizing Hypernetwork and sampling multiple samples in parallel during the learning process.
\begin{figure}[!t]
    \centering
    \includegraphics[width=1\linewidth]{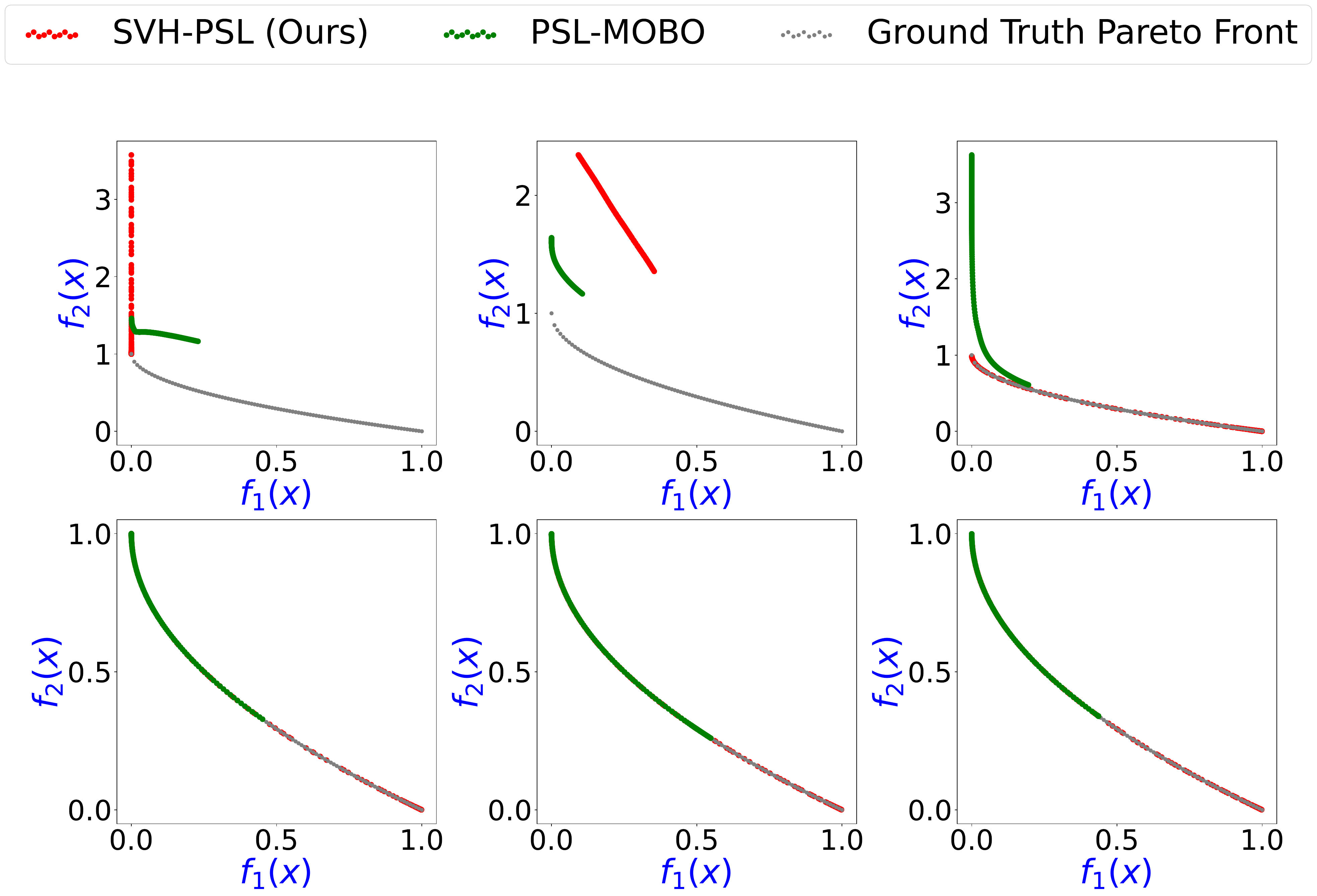}
    \caption{Approximate Pareto front comparison after the first 6 function evaluations using PSL-MOBO and our method, SVH-PSL, on the ZDT1 problem (2 objectives, 20 dimensions).}
    \label{fig:enter-label}
\end{figure}

\hl{Approach} In this paper, we propose a novel Pareto set learning model that simultaneously samples multiple solutions during the learning process. By leveraging the mutual interactions among these samples, our approach helps them escape local optima. These interactions also promote mutual repulsion, encouraging the samples to explore a diverse solution space. This results in a more diverse Pareto set, enhancing exploration while ensuring convergence to the optimal front. As shown in Figure \ref{fig:enter-label}, the Pareto front approximated by PSL-MOBO becomes stuck during the initial function evaluation steps, whereas our method, SVH-PSL, effectively overcomes this issue and successfully captures the entire Pareto front in the ZDT1 problem.

Building upon the previously established framework, we propose a novel methodology for multi-objective Bayesian optimization (MOBO) that employs Stein Variational Gradient Descent (SVGD) to enhance the learning of the Pareto set. This approach tackles challenges in expensive multi-objective optimization problems (EMOPs), where the evaluation of objective functions incurs high computational costs. Our contributions are as follows:
\begin{itemize}
    \item First, we present the mathematical formulation for controllable Pareto set learning in the context of expensive multi-objective optimization tasks.
    \item Second, we present SVH-PSL, an innovative framework for Pareto set learning in EMOPs, integrating Stein Variational Gradient Descent with a Hypernetwork and introducing a novel kernel for greater efficiency.
    \item Third, we perform extensive experiments on synthetic and real-world multi-objective optimization problems to assess the effectiveness of our proposed SVH-PSL method compared to baseline approaches.
\end{itemize}

\section{Preliminary}

\subsection{Expensive Multi-Objective Optimization}
We consider the following expensive continuous multi-objective optimization problem:

\begin{align}
    \label{eq: mop}
    \vx^* &= \argmin_{\vx \in \mathcal{X}} \ \vf(\vx), \\
    \vf(\vx) &= \big(f_1(\vx),f_2(\vx),\cdots, f_m(\vx)\big) \notag 
\end{align}    
where $\mathcal{X} \subset \mathbb{R}^n$ is the decision space, $f_i: \mathcal{X} \rightarrow \mathbb{R}$ is a black-box objective function. For a nontrivial problem, no single solution can optimize all objectives simultaneously, and there will always be a trade-off among them. We have the following definitions for multi-objective optimization: 

\begin{definition}[Dominance]
    A solution $\vx^a$ is said to dominate another solution $\vx^b$ if and only if $f_i(\vx^{a}) \leq f_i(\vx^b), \ \forall i \in \{1,...,m\}$ and $\vf(\vx^a) \neq \vf(\vx^b)$. We denote this relationship as $\vx^a \prec \vx^b$. 
\end{definition}

\begin{definition}[Pareto Optimality]
A solution $\vx^*$ is called Pareto optimal solution if $\nexists \vx^b \in \mathcal{X}: \ \vx^b \prec \vx^*$.  
\end{definition}

\begin{definition}[Pareto Set/Front]
The set of Pareto optimal is Pareto set, denoted by $\mathcal{P} = \{\vx^*\} \subseteq \mathcal X$ and the corresponding images in objectives space are Pareto front $\mathcal{P}_f = \{\vf(\vx) \mid \vx \in \mathcal{P}\}$.
\end{definition}

\begin{definition}[Hypervolume]
    Hypervolume \cite{zitzler1999multiobjective} is the area dominated by the Pareto front. Therefore, the quality of a Pareto front is proportional to its hypervolume. Given a set of $n$ points $\vy = \{y^{(i)} | y^{(i)} \in \mathbb{R}^m; i=1,\dots, n\}$ and a reference point $\rho \in\mathbb{R}^m$, the Hypervolume of $\vy$ is measured by the region of non-dominated points bounded above by $y^{(i)} \in \vy$, then the hypervolume metric is defined as follows:
    \begin{equation}
        HV(\vy) = VOL\left(\underset{y^{(i)} \in \vy, y^{(i)} \prec \rho}{\bigcup}\displaystyle{\Pi_{i=1}^n}\left[y^{(i)},\rho_i\right]\right)
    \end{equation}
 where $\rho_i$ is the i$^{th}$ coordinate of the reference point $\rho$ and $\Pi_{i=1}^n\left[y^{(i)},\rho_i\right]$ is the operator creating the n-dimensional hypercube from the ranges $\left[y^{(i)},\rho_i\right]$.
    
\end{definition}

\subsection{Gaussian Process and Bayesian Optimization }
A Gaussian Process with a single objective is characterized by a prior distribution defined over the function space as:
\begin{eqnarray}
f(\vx) \sim GP(\mu(\vx),k(\vx,\vx)),
\label{eq: gaussian_process}
\end{eqnarray}
where $\mu: \mathcal{X} \rightarrow \bbR$ represents the mean function, and $k: \mathcal{X} \times \mathcal{X} \rightarrow \bbR^2$ is the covariance kernel function. Given $n$ evaluated solutions $\vD = \{\vX, \vy\} = \{(\vx^{(i)},f(\vx^{(i)})|i = 1,\ldots,n)\}$,  the posterior distribution can be updated by maximizing the marginal likelihood based on the available data. For a new solution $\vx^{n+1}$, the posterior mean and variance are given by:
\begin{align*}
\hat{\mu}(\vx^{(n+1)}) &= \vk^T\vK^{-1}\vy, \\
\hat{\sigma}^2(\vx^{(n+1)}) &= k(\vx^{(n+1)},\vx^{(n+1)}) -  \vk^T \vK^{-1}\vk, 
\end{align*}
where $\vk = k(\vx^{(n+1)},\vX)$ is the kernel vector and $\vK = k(\vX, \vX)$ is the kernel matrix.

Bayesian optimization involves searching for the global optimum of a black-box function $f(\cdot)$ by wisely choosing the next evaluation via the current Gaussian process and acquisition functions. A new evaluation $\vx^{n+1}$ is determined through an acquisition function $\alpha$, which guides the search for the optimal solution. More specifically, the next evaluation $\vx^{n+1}$ is selected as the optimal solution of the acquisition in order to make the best improvement:
\begin{eqnarray}
\vx^{n+1} = \underset{x}{\argmax}\ \alpha(\vx; \vD),
\label{eq: acquisition_funcition}
\end{eqnarray}

 When a Gaussian process is used to approximate the unknown objective function, the acquisition function helps balance the trade-off between exploration (sampling areas with high uncertainty) and exploitation (sampling areas with promising results based on the model's predictions). There are many options available in the field of surrogate models, such as Expected Improvement (EI), Upper Confidence Bound (UCB), and Lower Confidence Bound (LCB). In the scope of this study, we choose to use LCB as our acquisition function, which is:
\begin{align}
    \hat{\vf}(\vx) = \hat{\mu}(\vx) -  \lambda \hat{\sigma}(\vx)
\end{align}

Additionally, in the setting of multi-objective optimization, hypervolume is a common criterion for determining the
quality of the Pareto front. For multi-objective Bayesian optimization, 
we use the concept of Hypervolume Improvement (HVI) is a valuable metric in multi-objective optimization that quantifies the increase in hypervolume achieved by adding the set of new solutions $\{ \vX_+ = \{\vx^{(i)}\}^b_{i=1}, \vY_+ = \hat{\vf}(\vX_+)  \}$ to a current set of solutions $\{\vX, \ \vY\}$:
\begin{equation}
    HVI\big(\vY_+,\ \vY\big) = HV\big( \vY_+ \cup \vY \big) - HV(\vY)
\end{equation}
Here $b$ represents the number of solutions. To select the optimal $\vX_+$ for the next evaluation, we choose those that maximize the HVI value. In other words, we use HVI as the additional acquisition function for selecting new solutions:
\begin{align}
    \vX_+ & = \argmax_{\mathbf{X}_+\in \vX} HVI(\vY_+, \vY) 
\end{align}

We select the set $\vX_+$ in a sequential greedy manner from $\vX$ where $|\vX|=1000$ is approximated by the Pareto set model, which we present in the section below.

\subsection{Pareto Set Learning}
\label{sec: PSL_EMOO}
Pareto Set Learning approximates the entire Pareto Front of Problem (\ref{eq: mop}) by directly approximating the mapping between an arbitrary preference vector $r$ and a corresponding Pareto optimal solution computed by surrogate models: 
\begin{align}
    \label{eq: PSL}
    & \theta^* = \argmin_{\theta} \mathbb{E}_{r \sim \text{Dir}(\alpha)} g(\hat{\vf}(\vx_r)| r) \\
    & \text{s.t } \ \vx_r = h(r|\ \theta) \in \mathcal{P}, \ h(\bbS^m|\ \theta^*) = \mathcal{P} \nonumber
\end{align}
where $\text{Dir}(\alpha)$ is the flat Dirichlet distribution with $\alpha = \left(\frac{1}{m}, \dots, \frac{1}{m} \right) \in \mathbb{R}^m$, $\bbS^m = \{ r \in \mathbb{R}^m_{>0}: \sum_i r_i = 1\}$ is the feasible space of preference vectors $r$, $\hat{f}_i(\cdot): \mathcal{X} \rightarrow \mathbb{R}, \forall i \in \{1,\dots,m\}$ are surrogate models, $\hat{\vf}(\cdot) = \left[ \hat{f}_i(\cdot) \right]_{i=1}^m$, the scalarization function  $g: \mathbb{R}^m \times \bbS^m \rightarrow \mathbb{R}$ helps us map a given preference vector with a Pareto solution, and $h (\cdot, \cdot): \bbS^m \times \Theta \rightarrow \mathcal{X}$ is called the Pareto set model, approximating the mentioned mapping.

The landscape provides various choices for scalarization functions like Linear Scalarization, Chebyshev, and Inverse Utility. However, we decided to use the Chebyshev function, 

\begin{align} \label{chebyshev}
    g(\hat{\vf}(\vx)| r) = \max_i \{r_i \lvert\hat{f}_i(\vx) - z_i^*\rvert \} \ \  \forall i \in \{1, \dots, m \}
\end{align}

where $z^*$ is an ideal objective vector. In black-box optimization, where the form of the objectives is unknown (and potentially non-convex), the Chebyshev scalarization is useful because it allows us to work with the problem in a simpler, scalarized form. Unlike traditional linear scalarization, which can struggle with non-convex Pareto fronts (as they may miss points on non-convex regions), Chebyshev scalarization can handle non-convex Pareto fronts more effectively by finding solutions across the entire front. 

In truth, obtaining a complete and accurate approximation of the entire variable space using surrogate models is impossible. However, the essence of Pareto set learning lies in the precise approximation of the feasible optimal space. This space, existing as a continuous manifold, represents a distinct subset within the broader variable space. Through a well-considered strategy, Pareto Set Learning demonstrates its ability to establish a highly accurate mapping between the preference vector space and the Pareto continuous manifold. However, optimizing the Pareto Set Model can be challenging and unstable if Gaussian processes do not approximate well black-box functions.

\subsection{Stein Variational Gradient Descent (SVGD)}
We present the Stein Variational Gradient Descent method, as proposed by \cite{liu2016stein}. SVGD serves as an effective approach for approximating intricate distributions by utilizing a collection of particles. These particles are iteratively updated to match
a target distribution more closely.

To approximate a given target distribution $p(x)$ by a set of particles $\{x_i\}_{i=1}^n$, we first draw a set of initial particles $\{x^0_i\}^n_{i=1}$ from the initial distribution $q(x)$, and then iteratively updating them with a deterministic transformation of form:
\begin{align}
    \vx_i \leftarrow \vx_i + \epsilon \phi^*_k(\vx_i), \forall \vi = 1,...,\vn, \\
    \phi^*_k = \arg\max_{\phi\ \in \mathcal{B}_k} \left\{ - \frac{d}{d\epsilon}\mathbf{K}\mathbf{L}(\vq_{[\epsilon \phi]} || \vp )\bigg|_{\epsilon=0}\right\}
\end{align}
here $\epsilon$ is a step size, $\phi^*$ is an optimal transform chosen to maximize the decreasing rate of the KL divergence between the distribution of particles the target $\vp$, and $\vq_{[\epsilon \phi]}$ is defined the distribution of the updated particles, and $\mathcal{B}_k$ is a unit ball of a reproducing kernel Hilbert space (RKHS) $\mathcal{H}^d_k := \mathcal{H}_k \times ... \times \mathcal{H}_k$, $\mathcal{H}_k$ is a Hilbert space associated with a positive definite kernel $\vk(\vx,\vx')$.

\cite{liu2016stein} refers to $\mathcal{P}$ as the Stein operator and shows that the optimal transform $\phi^*$ expressed:
\begin{equation}
\begin{split}
     & \phi^*_k \propto \mathbb{E}_{\vx \sim \vq}[\mathcal{P}\vk(x, \cdot )]  \\
    &= \mathbb{E}_{\vx \sim \vq} \left[\nabla_{\vx} \log\vp(\vx)\vk(\vx,\cdot)
    + \nabla_{\vx}\vk(\vx,\cdot) \right]
\end{split} 
\end{equation}
\section{Pareto Set Learning with SVGD}
This section outlines our primary framework. We present the fundamental concept of the alignment between Hypernetwork for sampling and the SVGD algorithm. This study introduces a novel methodology for acquiring the Pareto front by employing SVGD for gradient updates while incorporating a control factor through scalarization techniques. Next, we develop our local kernel to improve the efficiency of the SVGD optimization technique. 

The key idea is to observe the utilization of hypernetwork to generate a set of initial arbitrary particles. In the context of SVGD, choosing an optimal transform $\phi^*$ corresponds to the learning gradient to update the parameter of the hypernetwork. Our objective is to advance the particles toward the Pareto set.

\subsection{Stein Variational Hypernetworks} \label{sec: svgd-main}
The proposal involves employing Stein Variational Gradient Descent to adjust the gradient for the update direction in a method we refer to as Stein Variational Hypernetworks (SVH). This approach leverages the concept of Multi-Sample Hypernetwork \cite{hoang2023improving}, which involves sampling a set of preference vectors as input to the hypernetwork and generating a corresponding set of \(\vx_r\) as output.

We sample $K$ preference vectors $\vr_i$ and fit them to the Pareto set model $h(r|\theta)$ to generate solutions $\{\vx_{r_i}\}_{i=1}^K$. For a problem with $\vm$ objectives, the objective functions are represented as $\mathcal{F}_i(\vx_{r_i}) = [\hat{\vf_1}(\vx_{r_i}), ... , \hat{\vf_m}(\vx_{r_i})]$, where $\mathcal{F}_i$ is considered as a particle. The set of particles $\mathbf{F} = {\{\mathcal{F}_i}\}^K_{i=1}$ serves as the initialization set. We iteratively move the particles towards the Pareto front using the update rule:
\begin{equation}\label{updatephi}
\theta_{t+1} = \theta_t - \xi \nabla_\theta g \left( \mathcal{F}  \left( \vx_r \right)| r \right) 
\end{equation} 
where $\xi$ is the learning rate.

Stein Variational Gradient Descent plays a role in adjusting the direction of the gradient update and helps push the points away from each other, thereby enhancing the diversity of the Pareto front. Consequently, the following modifications will be applied to Formula \ref{updatephi}:
\begin{equation}\label{updatephi_svgd}
\theta_{t+1} = \theta_t - \xi \sum_{i=1}^K \sum_{j=1}^K \nabla_\theta g \left( \mathcal{F}_i | r_i \right) k(\mathcal{F}_i, \mathcal{F}_j) - \alpha\nabla_\theta k(\mathcal{F}_i, \mathcal{F}_j)
\end{equation} 
here, $k(\mathcal{F}_i, \mathcal{F}_j)$ is kernel matrix. Gaussian kernel is used: $k(\mathcal{F}_i, \mathcal{F}_j) = \exp(-\frac{1}{2c^2}||\mathcal{F}_i- \mathcal{F}_j||^2)$, $c$ is bandwidth, $\alpha$ is a positive coefficient that controls the importance of the divergence term.
\\
We can see in this formula that there are two main ideas:
\begin{itemize}
    \item  When considering $\nabla_\theta g\left(\mathcal{F}_i | r_i \right)$, it is our priority to guarantee that the gradient update direction aligns with our intended direction and is maintainable under control.
    \item The second term $\nabla_\theta k(\mathcal{F}_i, \mathcal{F}_j)$ plays a crucial role in the optimization process by facilitating the separation of particles. This term is primarily responsible for generating a repulsive force that pushes particles apart from each other \cite{liu2021profiling}.
\end{itemize}

Notably, our study proposes an innovative approach which is different from MOO-SVGD by \cite{liu2021profiling}. We employ the gradient of the scalarization function $\nabla_\theta g \left( \mathcal{F}_i | r_i \right)$ as a substitute for Multi-Gradient Descent Algorithm (MGDA) \cite{desideri2012multiple} in their work, which guarantees the convergence of the Pareto set learning method with preference vector, while simultaneously enhancing the diversity of the Pareto front in accordance with SVGD theory.

\subsection{Design of SVGD Local Kernel} \label{sec: lcoal-kernel}

The selection of the kernel significantly influences the efficacy of SVGD. The kernel employed in SVGD plays a pivotal role in determining the direction in which the particles are transformed, as it assigns weights to the contributions of each particle. Consequently, selecting an appropriate kernel is of utmost importance; while the Radial Basis Function (RBF) kernel utilizing a median heuristic is frequently adopted, it tends to be suboptimal for more challenging tasks. The concept of multiple kernel learning is presented by \cite{ai2023stein} under the designation MK-SVGD. This approach uses a composite kernel for approximation, with each kernel weighted by its importance.


In the context of expensive multi-objective optimization problems, surrogate models such as Gaussian Processes are often employed to approximate objective functions due to the high cost of direct evaluations. However, these models can lead to fragmented representations of the objective landscape, resulting in numerous pseudo-local optima—apparent optimal points that do not represent true global optima. Within our SVH-PSL framework, we generate a diverse set of particles using a hypernetwork, which can initially distribute across these various pseudo-local optima, which can lead to difficulties in particle movement and gradient updates. Traditional kernel computation considers the overall distance between points, which may not capture each dimension's influence in complex spaces.

To address this challenge, we introduce the concept of the local kernel for adapting to complex landscapes by calculating the kernel for each dimension separately, capturing its individual influence on particle movement.
\begin{align}
    \vk(\mathcal{F}_i, \mathcal{F}_j) = \sum^m_{n=1}\vk_n\left(\hat{\vf_n}(\vx_{r_i}), \hat{\vf_n}(\vx_{r_j}) \right).id_n
\end{align}
Here $\vk_n(\cdot)$ is the kernel for $n$ th dimension, $id_n$ is 1 in case $n$ is the index of the function $\hat{\vf_n}$ that achieves the maximum formula (\ref{chebyshev}), 0 otherwise.

\begin{figure}[!ht]
    \centering
    \includegraphics[width=0.9\linewidth]{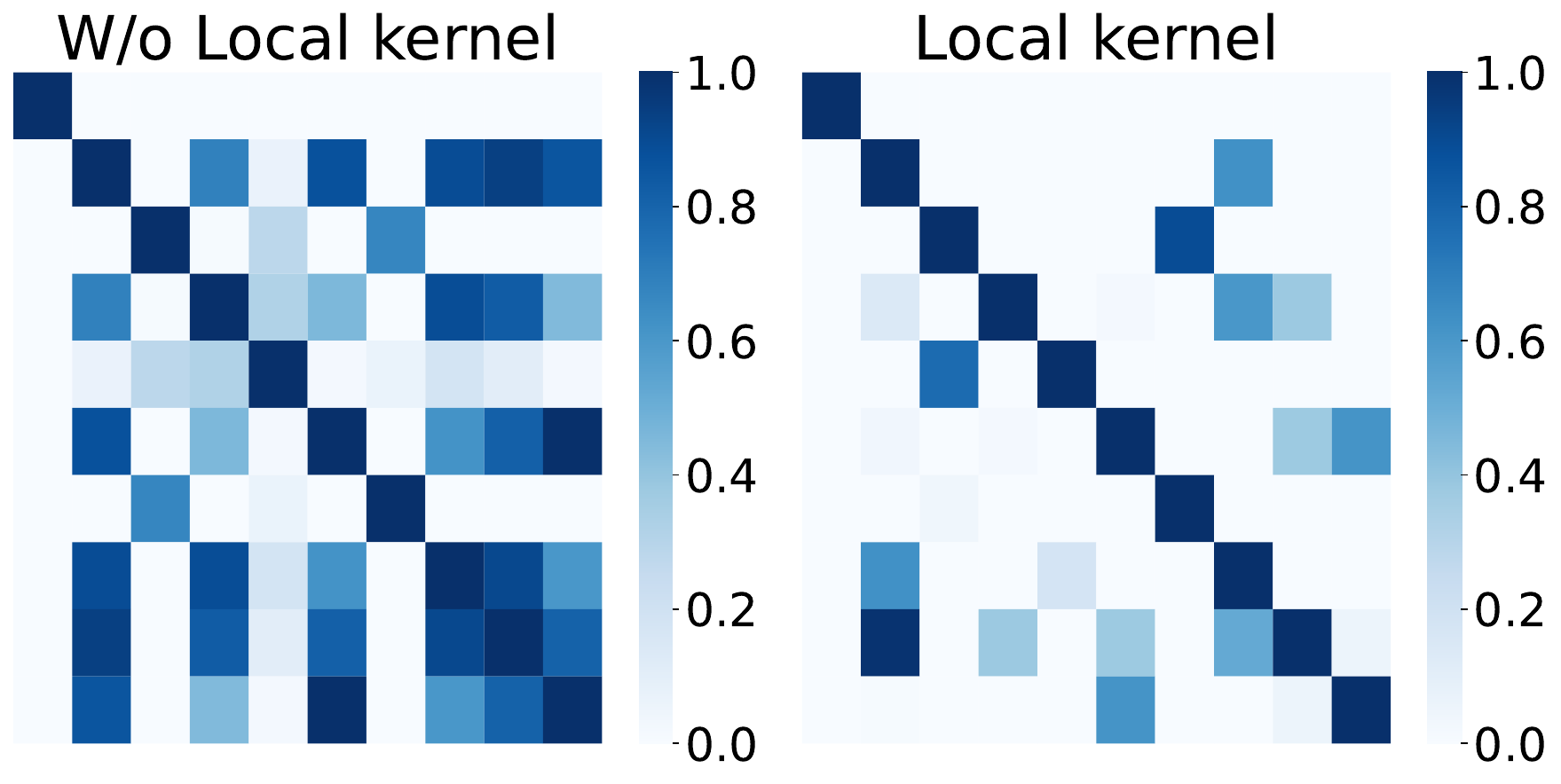}
    \caption{Comparison of the impact of neighboring particles on a particle with and without the local kernel. Here we are considering $K=10$}
    \label{fig:kernel}
\end{figure}

 This methodology facilitates a more detailed evaluation of particle interactions, which can help in overcoming the challenges posed by pseudo-local optima. By considering the influence of each dimension separately, we can achieve a more accurate modification of particle locations. With a more refined kernel computation, the gradient updates in SVGD can become more accurate, leading to improved convergence properties. Particles are less likely to get stuck in pseudo-local optima, as the dimensional influence can guide them more effectively towards globally optimal regions. Our method might enhance the robustness of SVGD by allowing it to adapt better to the geometry of the objective space, particularly when dealing with complex optimization problems.

 \hl{Discussion} Intuitively, SVH-PSL executes direct operations via the kernel to modify the gradient. It is well established that gradients are highly sensitive and can easily experience issues such as gradient explosion or vanishing. Figure \ref{fig:kernel} illustrates that when a particle is significantly affected by the presence of other particles, this interaction can facilitate a more rapid convergence. However, it may also lead to instability, potentially diverting the particle from reaching the global optimum. The implementation of our local kernel facilitates precise adjustments aimed at minimizing the potential for instabilities.

 \begin{algorithm}[ht!]
\caption{SVH-PSL main algorithm}
\label{algo: algorithm}
\textbf{Input}: Black-box multi-objective objectives $\vf(\vx) = \{f_j(\vx),\ j \in 1, \cdots, m\}$ and initial evaluation samples $\big\{\vx_0, \ \vf(\vx_0)\big\}$\\
\begin{algorithmic}[1] 
\STATE $\vD \la \big\{\vx_0, \ \vf(\vx_0)\big\} $
\STATE Initialize Pareto set model $h(r|\ \theta_0)$
\FOR{i $\la 0$ to $N$}
\STATE Training GP $\hat{f}_j(\vx)$ for each $f_j(\vx)$ on $\vD$
\FOR{t $\la 0$ to $T$}
\STATE Randomly sample $K$ vectors $\{r_i\}_{i=1}^K \sim \bbS^m$
\STATE Update $\theta_i$ with gradient descent by Formula (\ref{updatephi_svgd})
\ENDFOR
\STATE Randomly sample $B$ vectors $\{r_i\}_{i=1}^B \sim \bbS^m$, then compute $\vx_{r_i} = h(r_i|\ \theta)$, $\vX =\{\vx_{r_i}\}_{i=1}^B  $
\STATE Selecting subset $\{\vx\}_{b} \in \vX$ that has highest HVI
\STATE $\vD \la \vD \cup \big\{\vx, \ \vf(\vx)\big\}_{b}$
\ENDFOR
\end{algorithmic}
\textbf{Output}: Total evaluated solutions $\vD = \big\{\vx,\ \vf(\vx)\big\}$ and the final parameterized Pareto set model $h(r|\ \theta)$
\end{algorithm}
For each iteration, we trained the Pareto Set Model $h(r|\theta)$ with $T$ training step, whereas we randomly sampled $K$ preference $\{r_i\}_{i=1}^K \sim \text{Dir}(\alpha), \alpha \in \mathbb{R}^m$ for optimizing $\theta$. Then, we sampled $B$ preference $\{r_i\}_{i=1}^B \sim \text{Dir}(\alpha)$ to compute the corresponding evaluation $\vx_{r_i}$. Finally, we selected a subset $\{\vx\}_{b} \in \{\vx_{r_i}\}_{i=1}^B$ including $b$ elements that achieve the highest HVI for the next batch of expensive evaluations:
\begin{align}
    \{\vx\}_{b} & = \argmax_{\mathbf{X}_+} HVI(\hat\vf(\mathbf{x}_+), D_y) \\
    & \text{s.t } \left| \mathbf{x}_+ \right| = b, \ \mathbf{x}_+ \in \{\vx_{r_i}\}_{i=1}^B \notag
\end{align}

\section{Experiments}
{\bf Evaluation Metric.} We use Log Hypervolume Difference (LHD) to evaluate the quality of a learned Pareto front, denoted as $\mathcal{P}_f$, compared to the true approximate Pareto front for the synthetic/real-world problems, represented as $\hat{\mathcal{P}}_f$. The calculation of LHD involves taking the logarithm of the difference in hypervolumes between $\hat{\mathcal{P}}_f$ and $\mathcal{P}_f$ as follows:
\begin{equation*}
LHD(\mathcal{P}_f, \hat{\mathcal{P}}_f) = \log\Big(\text{HV}\big(\mathcal{P}_f\big) - \text{HV}\big(\hat{\mathcal{P}}_f\big)\Big)    
\end{equation*}
\begin{figure*}[ht!]
    \centering
     \includegraphics[scale =0.159]{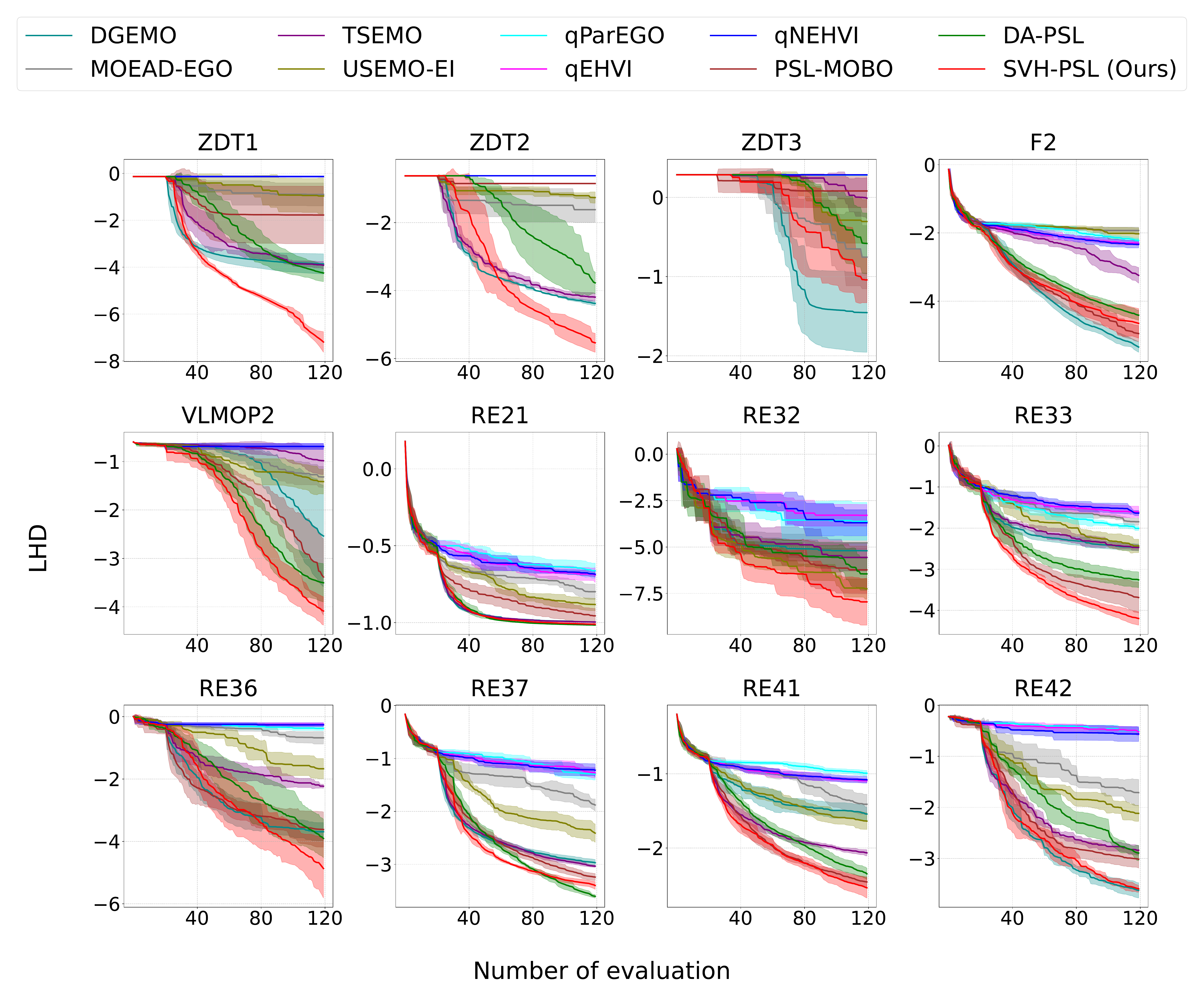}
    \caption{Mean Log Hypervolume Differences between the truth Pareto Front and the learned Pareto Front with respect to the number of expensive evaluations on all MOBO algorithms.}
    \label{fig: comparision_LHD}
\end{figure*}
\\{\bf Baseline Methods.} We compare SVH-PSL with the current state-of-the-art MOBO PFL methods, including TS-TCH \cite{paria2020flexible}, USeMO-EI \cite{belakaria2020uncertainty}, MOEA/D-EGO \cite{zhang2009expensive}, TSEMO \cite{bradford2018efficient}, DGEMO \cite{konakovic2020diversity}, qParEGO \cite{1583627}, qEHVI \cite{daulton2020differentiable}, qNEHVI \cite{daulton2021parallel}, DA-PSL \cite{lu2024you} and PSL-MOBO \cite{lin2022pareto}. 
\\
{\bf Synthetic and Real-World Benchmarks.} To demonstrate the effectiveness of our proposed method, we perform experiments in synthetic benchmarks and real-world application datasets; each dataset contains 2 or 3 constraint objectives. For the synthetic test, we choose 5 problems in the class of problem ZDT \cite{deb2006innovization} and F2, VLMOP2 \cite{van1999multiobjective} with 2 objectives. For the real-world issues, we test our hypothesis in 7 over 16 datasets summarized by Tanabe and Ishibuchi \cite{tanabe2020easy}. In our experiments, RE problems are denoted as RExy, where x is the number of objectives and y is the id of the problems, ie, the RE21 means the real-world problems with 2 objectives and has id 1, associated with the issues RE2-4-1. The same definition applies to RE32, RE33, RE36, RE37, RE41, and RE42. 
\subsection{Experiment Settings}
For all experiments\footnote{\scriptsize Code is available at \url{https://github.com/nguyenduc810/SVH-PSL}}, we randomly generate 20 initial solutions for expensive evaluations. The model is trained over 20 iterations with a batch size of b = 5. In Formula \ref{updatephi_svgd}, the hyperparameters are set as $c=med$ (the median of the pairwise distances between samples) and $\alpha=0.1$. Each problem is trained 5 times to eliminate randomness. 
\subsection{Experimental Results and Analysis}
\begin{figure}[!ht]
    \centering
    \includegraphics[width = .468\textwidth]{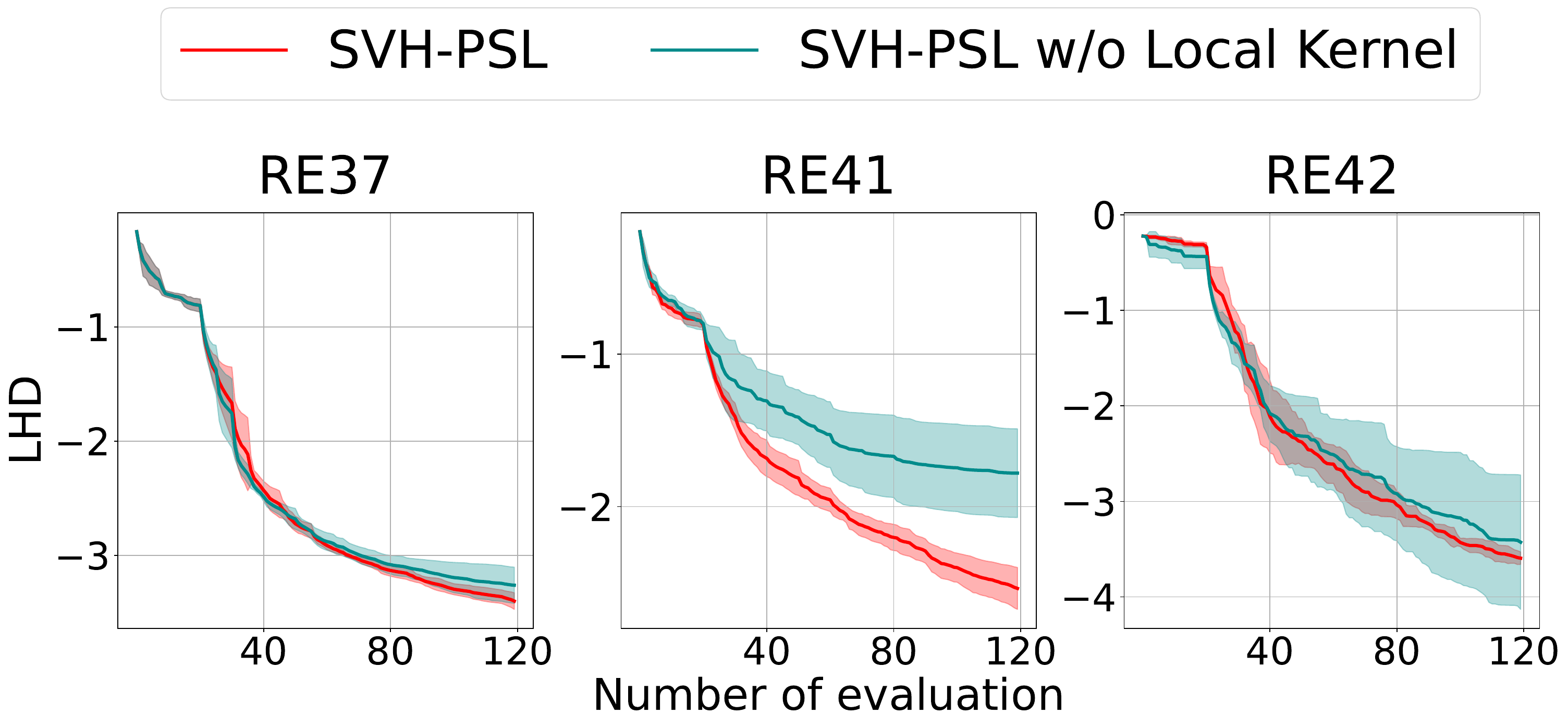}
    \caption{
    Illustration of SVH-PSL performance with and without local kernel integration in real-world problems with a complex PF.}
    \label{fig:compare_ker}
\end{figure}

\hl{MOBO Performance} Figure \ref{fig: comparision_LHD} compares the log hypervolume difference (LHD) of baseline methods, with solid lines showing the mean and shaded regions indicating standard deviation. Our proposed SVH-PSL demonstrates superior performance, converging rapidly in synthetic experiments and achieving strong results in real-world scenarios. Notably, the integration of the SVGD algorithm enhances convergence speed compared to the conventional PSL-MOBO method while also providing better diversity on the Pareto set than the DA-PSL method.
\\
\hl{Analysis of Local Kernel Benefits in Complex Pareto Fronts} Figure \ref{fig:compare_ker} demonstrates the effectiveness of local kernel integration in SVH-PSL, especially in solving real-world problems with complex Pareto fronts (RE37, RE41, and RE42). SVH-PSL achieves a consistently lower LHD compared to SVH-PSL without the local kernel. Moreover, SVH-PSL with the local kernel exhibits a smaller variance, indicating that the local kernel adapts effectively to the complex Pareto front. This results in more stable and consistent solutions across multiple evaluations.
\subsection{Ablation Studies}
\begin{figure}[!ht]
    \centering
    \includegraphics[width = .37\textwidth]{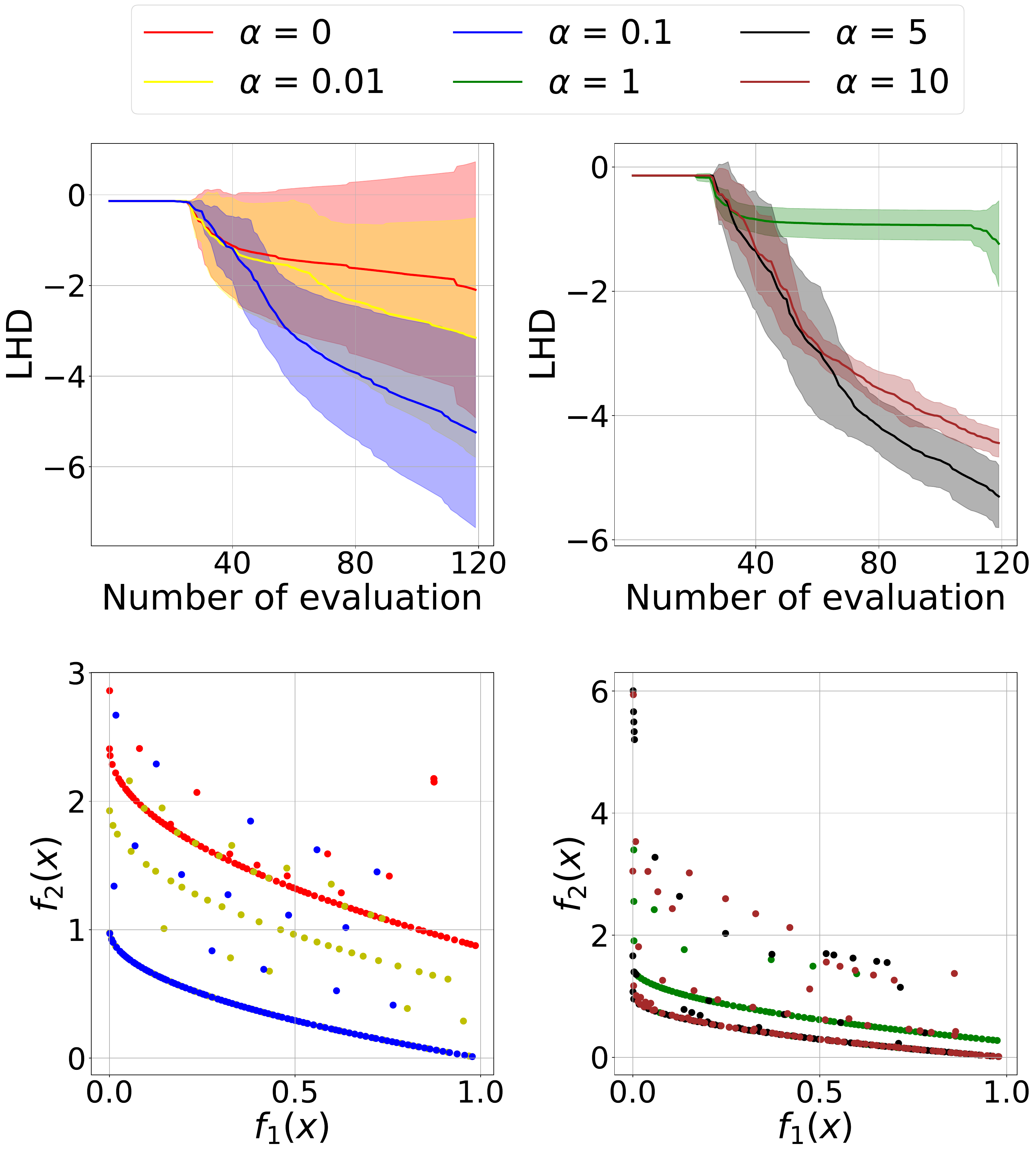}
    \caption{ Impact of $\alpha$ on the trade-off between exploration and exploitation in ZDT1 (LHD \& Pareto front).}
    \label{fig:ablation_alpha}
\end{figure}

\begin{figure}[!ht]
    \centering
    \includegraphics[width = .395\textwidth]{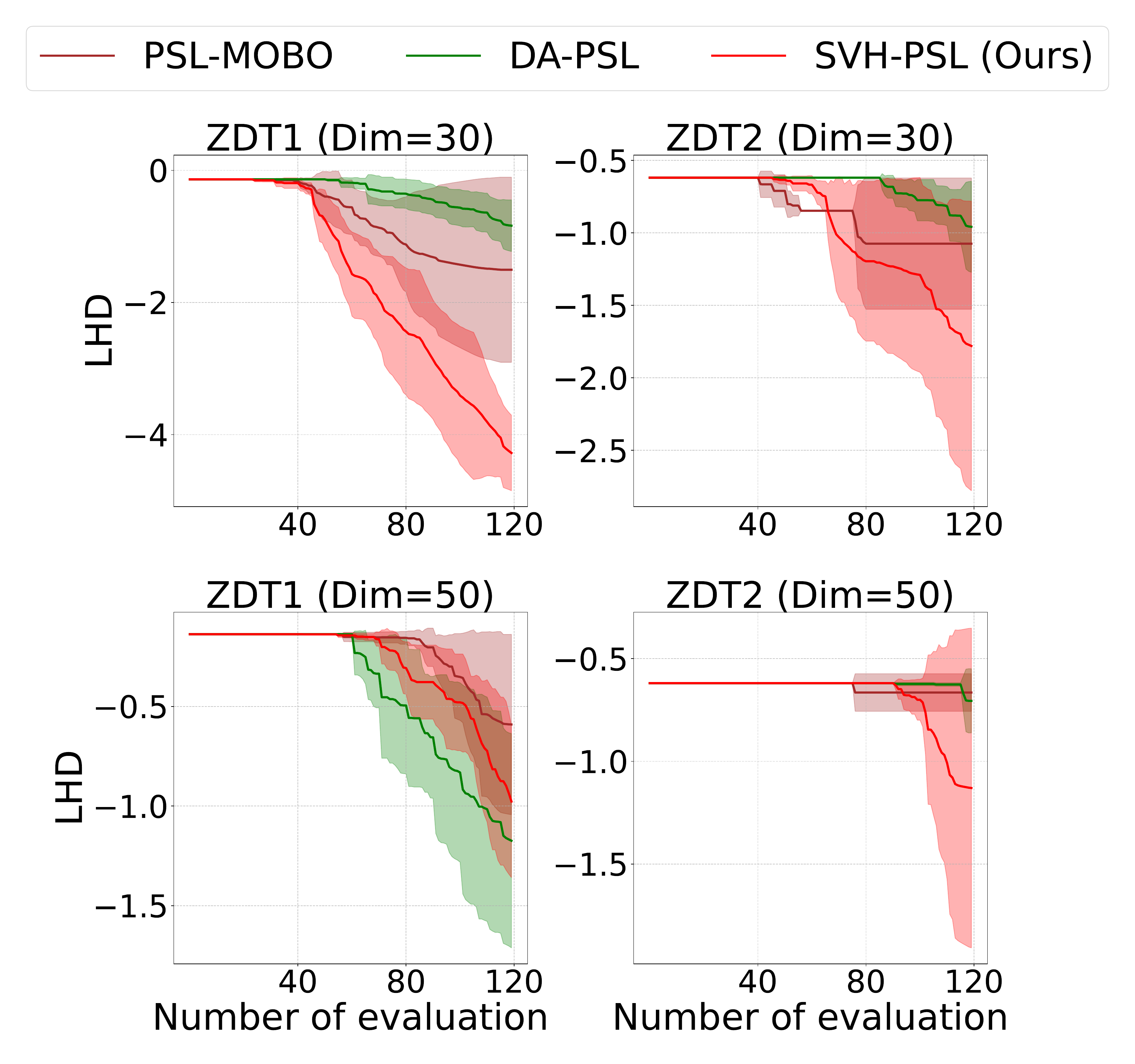}
    \caption{
    Performance comparison of high-dimensional ZDT1 and ZDT2 problems.}
    \label{fig:compare_dim}
\end{figure}

\hl{The Trade-Off Between Exploration and Exploitation}
In Formula \ref{updatephi_svgd}, the second term acts as the repulsion component, with the parameter $\alpha$ serving as a trade-off between exploration and exploitation. As shown in Figure \ref{fig:ablation_alpha}, $\alpha$ significantly impacts performance. When $\alpha$ is small (left), the variance is high, indicating that insufficient repulsion makes it difficult for the particles to escape pseudo-optimal points, leading to instability. Conversely, a larger $\alpha$ (right) encourages exploration, driving the particles toward the optimal solution.
\hl{High Dimensional} Figure \ref{fig:compare_dim} shows that SVH-PSL performs well in high-dimensional conditions. Although optimization problems become increasingly challenging as dimensionality rises, SVH-PSL maintains its effectiveness. However, it also exhibits higher variance.

\section{Related Work}
\label{sec: related}
{\bf Multi-objective Bayesian Optimization.} 
Conventional Multi-objective Bayesian optimization (MOBO) methods have primarily concentrated on locating singular or limited sets of solutions. To achieve a diverse array of solutions catering to varied preferences, scalarization functions have emerged as a prevalent approach. Notably, \citet{paria2020flexible} adopts a strategy of scalarizing the Multi-objective problem into a series of single-objective ones, integrating random preference vectors during optimization to yield a collection of diverse solutions. Meanwhile, \citet{abdolshah2019multiobjective} delves into distinct regions on the Pareto Front through preference-order constraints, grounded in the Pareto Stationary equation.

Alternatively, evolutionary and genetic algorithms have also played a role in furnishing diversified solution sets, as observed in \citet{zhang2009expensive} which concurrently tackles a range of surrogate scalarized subproblems within the MOEA/D framework \cite{zhang2007moea}. Additionally,  \citet{bradford2018efficient} adeptly combines Thompson Sampling with Hypervolume Improvement to facilitate the selection of successive candidates. Complementing these approaches, \citet{konakovic2020diversity} employs a well-crafted local search strategy coupled with a specialized mechanism to actively encourage the exploration of diverse solutions. A notable departure from conventional methods, \citet{belakaria2020uncertainty} introduces an innovative uncertainty-aware search framework, orchestrating the optimization of input selection through surrogate models. This framework efficiently addresses MOBO problems by discerning and scrutinizing promising candidates based on measures of uncertainty.

{\bf Pareto Set Learning}
\label{sec: related_PSL}
or PSL emerges as a proficient means of approximating the complete Pareto front, the set of optimal solutions in multi-objective optimization (MOO) problems, employing Hypernetworks \cite{chauhan2023brief}. This involves the utilization of hypernetworks to estimate the relationship between arbitrary preference vectors and their corresponding Pareto optimal solutions. In the realm of known functions, effective solutions for MOO problems have been demonstrated by \citet{navon2021learning, hoang2023improving, tuan2024framework} while in the context of combinatorial optimization, \cite{lin2022pareto_com} have made notable contributions. Notably, \citet{lin2022pareto} introduced a pioneering approach to employing PSL in the optimization of multiple black-box functions, PSL-MOBO, leveraging surrogate models to learn preference mapping, making it the pioneer method in exploring Pareto Set learning for the black-box multi-objective optimization task. However, PSL-MOBO is hindered by its reliance on Gaussian processes for training the Pareto set model, leading to issues of inadequacy and instability in learning the Pareto front.

{\bf Stein Variational Gradient Descent} is a particle-based inference method introduced by \cite{liu2016stein}. This approach has gained significant attention due to its ability to approximate complex probability distributions through deterministic sampling methods, making it a powerful alternative to traditional Markov Chain Monte Carlo (MCMC) techniques \cite{neal2012mcmc, hoffman2014no}. Recent research has utilized SVGD in the context of multi-objective optimization (MOO) challenges \cite{liu2021profiling,phan2022stochastic}.
\section{Conclusion}
This study presents a novel methodology for learning the Pareto set, utilizing Stein Variational Gradient Descent in conjunction with a Hypernetwork. This framework proves to be effective in approximating the complete Pareto Front while simultaneously optimizing several conflicting black-box objectives. The findings from our experiments further indicate that the method effectively acquires a diverse Pareto set, which facilitates robust exploration and exploitation. 
\section{Acknowledgments}
This research was funded by Vingroup Innovation Foundation (VINIF) under project code VINIF.2024.DA113.
\small
\bibliography{aaai25}

\begin{thebibliography}{37}
\providecommand{\natexlab}[1]{#1}

\bibitem[{Abdolshah et~al.(2019)Abdolshah, Shilton, Rana, Gupta, and Venkatesh}]{abdolshah2019multiobjective}
Abdolshah, M.; Shilton, A.; Rana, S.; Gupta, S.; and Venkatesh, S. 2019.
\newblock Multi-objective Bayesian optimisation with preferences over objectives.
\newblock arXiv:1902.04228.

\bibitem[{Ai et~al.(2023)Ai, Liu, He, and Xu}]{ai2023stein}
Ai, Q.; Liu, S.; He, L.; and Xu, Z. 2023.
\newblock Stein variational gradient descent with multiple kernels.
\newblock \emph{Cognitive Computation}, 15(2): 672--682.

\bibitem[{Attia et~al.(2020)Attia, Grover, Jin, Severson, Markov, Liao, Chen, Cheong, Perkins, Yang, Herring, Aykol, Harris, Braatz, Ermon, and Chueh}]{Attia2020ClosedloopOO}
Attia, P.~M.; Grover, A.; Jin, N.; Severson, K.~A.; Markov, T.~M.; Liao, Y.-H.; Chen, M.~H.; Cheong, B.; Perkins, N.; Yang, Z.; Herring, P.~K.; Aykol, M.; Harris, S.~J.; Braatz, R.~D.; Ermon, S.; and Chueh, W.~C. 2020.
\newblock Closed-loop optimization of fast-charging protocols for batteries with machine learning.
\newblock \emph{Nature}, 578: 397--402.

\bibitem[{Belakaria et~al.(2020)Belakaria, Deshwal, Jayakodi, and Doppa}]{belakaria2020uncertainty}
Belakaria, S.; Deshwal, A.; Jayakodi, N.~K.; and Doppa, J.~R. 2020.
\newblock Uncertainty-aware search framework for multi-objective Bayesian optimization.
\newblock In \emph{Proceedings of the AAAI Conference on Artificial Intelligence}, volume~34, 10044--10052.

\bibitem[{Bradford, Schweidtmann, and Lapkin(2018)}]{bradford2018efficient}
Bradford, E.; Schweidtmann, A.~M.; and Lapkin, A. 2018.
\newblock Efficient multiobjective optimization employing Gaussian processes, spectral sampling and a genetic algorithm.
\newblock \emph{Journal of global optimization}, 71(2): 407--438.

\bibitem[{Chauhan et~al.(2023)Chauhan, Zhou, Lu, Molaei, and Clifton}]{chauhan2023brief}
Chauhan, V.~K.; Zhou, J.; Lu, P.; Molaei, S.; and Clifton, D.~A. 2023.
\newblock A Brief Review of Hypernetworks in Deep Learning.
\newblock \emph{arXiv preprint arXiv:2306.06955}.

\bibitem[{Daulton, Balandat, and Bakshy(2020)}]{daulton2020differentiable}
Daulton, S.; Balandat, M.; and Bakshy, E. 2020.
\newblock Differentiable expected hypervolume improvement for parallel multi-objective Bayesian optimization.
\newblock \emph{Advances in Neural Information Processing Systems}, 33: 9851--9864.

\bibitem[{Daulton, Balandat, and Bakshy(2021)}]{daulton2021parallel}
Daulton, S.; Balandat, M.; and Bakshy, E. 2021.
\newblock Parallel bayesian optimization of multiple noisy objectives with expected hypervolume improvement.
\newblock \emph{Advances in Neural Information Processing Systems}, 34: 2187--2200.

\bibitem[{Deb and Srinivasan(2006)}]{deb2006innovization}
Deb, K.; and Srinivasan, A. 2006.
\newblock Innovization: Innovating design principles through optimization.
\newblock In \emph{Proceedings of the 8th annual conference on Genetic and evolutionary computation}, 1629--1636.

\bibitem[{D{\'e}sid{\'e}ri(2012)}]{desideri2012multiple}
D{\'e}sid{\'e}ri, J.-A. 2012.
\newblock Multiple-gradient descent algorithm (MGDA) for multiobjective optimization.
\newblock \emph{Comptes Rendus Mathematique}, 350(5-6): 313--318.

\bibitem[{He et~al.(2023)He, Zhang, Gong, and Ji}]{he2023review}
He, C.; Zhang, Y.; Gong, D.; and Ji, X. 2023.
\newblock A review of surrogate-assisted evolutionary algorithms for expensive optimization problems.
\newblock \emph{Expert Systems with Applications}, 217: 119495.

\bibitem[{Hoang et~al.(2023)Hoang, Le, Tuan, and Thang}]{hoang2023improving}
Hoang, L.~P.; Le, D.~D.; Tuan, T.~A.; and Thang, T.~N. 2023.
\newblock Improving pareto front learning via multi-sample hypernetworks.
\newblock In \emph{Proceedings of the AAAI Conference on Artificial Intelligence}, volume~37, 7875--7883.

\bibitem[{Hoffman, Gelman et~al.(2014)}]{hoffman2014no}
Hoffman, M.~D.; Gelman, A.; et~al. 2014.
\newblock The No-U-Turn sampler: adaptively setting path lengths in Hamiltonian Monte Carlo.
\newblock \emph{J. Mach. Learn. Res.}, 15(1): 1593--1623.

\bibitem[{Hou, Chen, and Pei(2024)}]{hou2024optimization}
Hou, M.; Chen, F.; and Pei, Y. 2024.
\newblock Optimization of geometric parameters of ejector for fuel cell system based on multi-objective optimization method.
\newblock \emph{International Journal of Green Energy}, 21(2): 228--243.

\bibitem[{Knowles(2006)}]{1583627}
Knowles, J. 2006.
\newblock ParEGO: a hybrid algorithm with on-line landscape approximation for expensive multiobjective optimization problems.
\newblock \emph{IEEE Transactions on Evolutionary Computation}, 10(1): 50--66.

\bibitem[{Konakovic~Lukovic, Tian, and Matusik(2020)}]{konakovic2020diversity}
Konakovic~Lukovic, M.; Tian, Y.; and Matusik, W. 2020.
\newblock Diversity-guided multi-objective bayesian optimization with batch evaluations.
\newblock \emph{Advances in Neural Information Processing Systems}, 33: 17708--17720.

\bibitem[{Laumanns and Ocenasek(2002)}]{laumanns2002bayesian}
Laumanns, M.; and Ocenasek, J. 2002.
\newblock Bayesian optimization algorithms for multi-objective optimization.
\newblock In \emph{International Conference on Parallel Problem Solving from Nature}, 298--307. Springer.

\bibitem[{Le and Lauw(2017)}]{le2017indexable}
Le, D.~D.; and Lauw, H.~W. 2017.
\newblock Indexable bayesian personalized ranking for efficient top-k recommendation.
\newblock In \emph{Proceedings of the 2017 ACM on Conference on Information and Knowledge Management}, 1389--1398.

\bibitem[{Lee et~al.(2024)Lee, Li, Ke, Yoo, Zhang, Yu, Wang, Deng, Entis, He et~al.}]{lee2024parrot}
Lee, S.~H.; Li, Y.; Ke, J.; Yoo, I.; Zhang, H.; Yu, J.; Wang, Q.; Deng, F.; Entis, G.; He, J.; et~al. 2024.
\newblock Parrot: Pareto-optimal Multi-Reward Reinforcement Learning Framework for Text-to-Image Generation.
\newblock \emph{arXiv preprint arXiv:2401.05675}.

\bibitem[{Lin, Yang, and Zhang(2022)}]{lin2022pareto_com}
Lin, X.; Yang, Z.; and Zhang, Q. 2022.
\newblock Pareto Set Learning for Neural Multi-Objective Combinatorial Optimization.
\newblock In \emph{International Conference on Learning Representations}.

\bibitem[{Lin et~al.(2022)Lin, Yang, Zhang, and Zhang}]{lin2022pareto}
Lin, X.; Yang, Z.; Zhang, X.; and Zhang, Q. 2022.
\newblock Pareto set learning for expensive multi-objective optimization.
\newblock \emph{Advances in Neural Information Processing Systems}, 35: 19231--19247.

\bibitem[{Liu and Wang(2016)}]{liu2016stein}
Liu, Q.; and Wang, D. 2016.
\newblock Stein variational gradient descent: A general purpose bayesian inference algorithm.
\newblock \emph{Advances in neural information processing systems}, 29.

\bibitem[{Liu, Tong, and Liu(2021)}]{liu2021profiling}
Liu, X.; Tong, X.; and Liu, Q. 2021.
\newblock Profiling pareto front with multi-objective stein variational gradient descent.
\newblock \emph{Advances in Neural Information Processing Systems}, 34: 14721--14733.

\bibitem[{Lu, Li, and Zhou(2024)}]{lu2024you}
Lu, Y.; Li, B.; and Zhou, A. 2024.
\newblock Are You Concerned about Limited Function Evaluations: Data-Augmented Pareto Set Learning for Expensive Multi-Objective Optimization.
\newblock In \emph{Proceedings of the AAAI Conference on Artificial Intelligence}, volume~38, 14202--14210.

\bibitem[{Milojkovic et~al.(2019)Milojkovic, Antognini, Bergamin, Faltings, and Musat}]{milojkovic2019multi}
Milojkovic, N.; Antognini, D.; Bergamin, G.; Faltings, B.; and Musat, C. 2019.
\newblock Multi-gradient descent for multi-objective recommender systems.
\newblock \emph{arXiv preprint arXiv:2001.00846}.

\bibitem[{Navon et~al.(2021)Navon, Shamsian, Fetaya, and Chechik}]{navon2021learning}
Navon, A.; Shamsian, A.; Fetaya, E.; and Chechik, G. 2021.
\newblock Learning the Pareto Front with Hypernetworks.
\newblock In \emph{International Conference on Learning Representations}.

\bibitem[{Neal(2012)}]{neal2012mcmc}
Neal, R.~M. 2012.
\newblock MCMC using Hamiltonian dynamics.
\newblock \emph{arXiv preprint arXiv:1206.1901}.

\bibitem[{Paria, Kandasamy, and P{\'o}czos(2020)}]{paria2020flexible}
Paria, B.; Kandasamy, K.; and P{\'o}czos, B. 2020.
\newblock A flexible framework for multi-objective bayesian optimization using random scalarizations.
\newblock In \emph{Uncertainty in Artificial Intelligence}, 766--776. PMLR.

\bibitem[{Phan et~al.(2022)Phan, Tran, Le, Tran, Ho, and Phung}]{phan2022stochastic}
Phan, H.; Tran, N.; Le, T.; Tran, T.; Ho, N.; and Phung, D. 2022.
\newblock Stochastic multiple target sampling gradient descent.
\newblock \emph{Advances in neural information processing systems}, 35: 22643--22655.

\bibitem[{Schweikard et~al.(2000)Schweikard, Glosser, Bodduluri, Murphy, and Adler}]{Schweikard2000RoboticMC}
Schweikard, A.; Glosser, G.~D.; Bodduluri, M.; Murphy, M.~J.; and Adler, J.~R. 2000.
\newblock Robotic motion compensation for respiratory movement during radiosurgery.
\newblock \emph{Computer aided surgery : official journal of the International Society for Computer Aided Surgery}, 5 4: 263--77.

\bibitem[{Swersky, Snoek, and Adams(2013)}]{swersky2013multi}
Swersky, K.; Snoek, J.; and Adams, R.~P. 2013.
\newblock Multi-task bayesian optimization.
\newblock \emph{Advances in neural information processing systems}, 26.

\bibitem[{Tanabe and Ishibuchi(2020)}]{tanabe2020easy}
Tanabe, R.; and Ishibuchi, H. 2020.
\newblock An easy-to-use real-world multi-objective optimization problem suite.
\newblock \emph{Applied Soft Computing}, 89: 106078.

\bibitem[{Tuan et~al.(2024)Tuan, Hoang, Le, and Thang}]{tuan2024framework}
Tuan, T.~A.; Hoang, L.~P.; Le, D.~D.; and Thang, T.~N. 2024.
\newblock A framework for controllable pareto front learning with completed scalarization functions and its applications.
\newblock \emph{Neural Networks}, 169: 257--273.

\bibitem[{Van~Veldhuizen and Lamont(1999)}]{van1999multiobjective}
Van~Veldhuizen, D.~A.; and Lamont, G.~B. 1999.
\newblock Multiobjective evolutionary algorithm test suites.
\newblock In \emph{Proceedings of the 1999 ACM symposium on Applied computing}, 351--357.

\bibitem[{Zhang and Li(2007)}]{zhang2007moea}
Zhang, Q.; and Li, H. 2007.
\newblock MOEA/D: A multiobjective evolutionary algorithm based on decomposition.
\newblock \emph{IEEE Transactions on evolutionary computation}, 11(6): 712--731.

\bibitem[{Zhang et~al.(2009)Zhang, Liu, Tsang, and Virginas}]{zhang2009expensive}
Zhang, Q.; Liu, W.; Tsang, E.; and Virginas, B. 2009.
\newblock Expensive multiobjective optimization by MOEA/D with Gaussian process model.
\newblock \emph{IEEE Transactions on Evolutionary Computation}, 14(3): 456--474.

\bibitem[{Zitzler and Thiele(1999)}]{zitzler1999multiobjective}
Zitzler, E.; and Thiele, L. 1999.
\newblock Multiobjective evolutionary algorithms: a comparative case study and the strength Pareto approach.
\newblock \emph{IEEE transactions on Evolutionary Computation}, 3(4): 257--271.

\end{thebibliography}

\end{document}